\documentclass[10pt,twocolumn,letterpaper]{article}

\usepackage[pagenumbers]{cvpr} 

\definecolor{cvprblue}{rgb}{0.21,0.49,0.74}
\usepackage[pagebackref,breaklinks,colorlinks,allcolors=cvprblue]{hyperref}
\usepackage{amsmath,amssymb}
\usepackage{graphicx}
\usepackage{array}
\usepackage{xcolor}
\usepackage{hyperref}
\newcommand{\bluelight}[1]{\textcolor{blue}{#1}}

\def\confName{CVPR}
\def\confYear{2027}

\title{3AGameFactory-Motion: Agentic Rigging and Reliable Complex Motion Generation for Diverse 3D Assets}

\author{
{\normalsize Bohan Zeng$^{1}$\quad Kaixin Zhu$^{2}$\quad Yifan Yang$^{3}$\quad Dongge Yu$^{1}$\quad Zihao Wu$^{1}$\quad Yifan Dai$^{4}$}\\
{\normalsize Zhou Liu$^{1}$\quad Junbo Niu$^{1}$\quad Xiaochen Ma$^{5}$\quad Shiran Ge$^{6}$\quad Zimo Meng$^{1}$\quad Junzhu Liang$^{1}$}\\
{\normalsize Huanyao Zhang$^{1}$\quad Jinhang Liu$^{1}$\quad Tianqi Xiao$^{1}$\quad Jinhao Zhang$^{7}$\quad Tianyi Bai$^{7}$\quad Zeno Zeng$^{7}$}\\
{\normalsize Penghao Zhao$^{7}$\quad Yuanbo Peng$^{7}$\quad Key Zhang$^{7}$\quad Wentao Zhang$^{1}$}\\[8pt]
{\small $^{1}$Peking University\quad $^{2}$Nanyang Technological University\quad $^{3}$Northeastern University}\\
{\small $^{4}$Shanghai Jiao Tong University\quad $^{5}$The Hong Kong University of Science and Technology}\\
{\small $^{6}$Institute of Automation, Chinese Academy of Sciences\quad $^{7}$Tencent Hunyuan}\\[6pt]
{\tt\small bhzeng25@stu.pku.edu.cn, wentao.zhang@pku.edu.cn}\\
{\tt\small Code: \texorpdfstring{\url{https://github.com/OpenDCAI/GameFactory-3A}}{https://github.com/OpenDCAI/GameFactory-3A}}
}

\makeatletter
\let\OriginalMakeTitle\@maketitle
\renewcommand{\@maketitle}{%
    \OriginalMakeTitle
    \begin{center}
        \includegraphics[width=\textwidth]{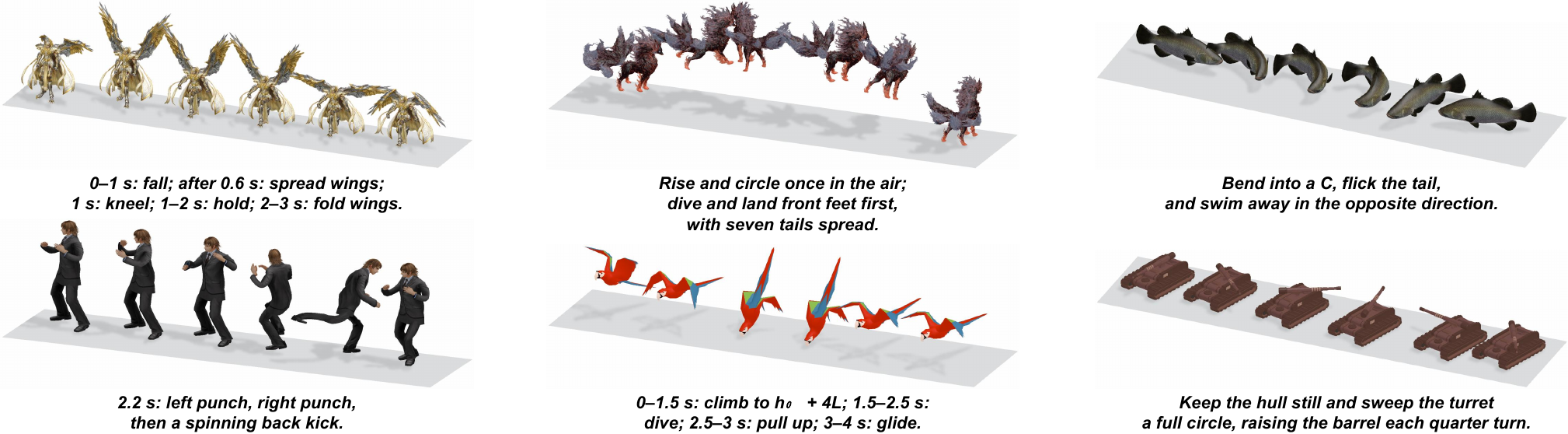}\par
        \vspace{0.5em}
        \refstepcounter{figure}\label{fig:teaser}
        {\small\textbf{Figure~\thefigure.} Overview of rigging and instruction-driven animation for diverse 3D assets.\par}
    \end{center}
    \vspace{1em}
}
\makeatother

\begin{document}
\maketitle
\begin{abstract}
Animating arbitrary 3D assets from instructions requires both structural inference and motion control, yet existing motion benchmarks mostly cover conventional human or animal skeletons with simple instructions. We introduce DivDynBench, which evaluates rigging and motion generation across six morphology categories, with instructions requiring multi-phase part coordination and explicit timing, angle, repetition, and trajectory constraints. With large language models, we systematically compare three common paradigms: specialized motion generation, video generation followed by motion recovery, and direct agent-driven generation. We analyze their control capabilities, cross-morphology adaptability, and generation costs. Guided by this comparison, we propose an Agentic Motion Framework in which an agent writes compact parametric skeleton and motion programs, and a deterministic executor performs rigging, kinematic solving, and geometric correction. Compact parameter control reduces generation overhead and improves rigging reliability and motion-control accuracy, providing practical guidance for deploying complex 3D motion generation.
\end{abstract}
    
\section{Introduction}

\begin{table*}[t]
\centering
\footnotesize
\setlength{\tabcolsep}{3pt}
\renewcommand{\arraystretch}{1.3}
\newcommand{\tabgood}{\textcolor{green!55!black}{\ensuremath{\checkmark}}}
\newcommand{\tabmid}{\textcolor{blue!65!black}{\ensuremath{\triangle}}}
\newcommand{\tabbad}{\textcolor{red!75!black}{\ensuremath{\times}}}
\caption{\centering Three LLM-based paradigms for complex 3D motion generation. \tabgood, \tabmid, and \tabbad indicate a relative strength, partial support, and a key limitation, respectively.}
\label{tab:intro_paradigms}
\begin{tabular}{@{}p{0.14\textwidth}|p{0.20\textwidth}p{0.24\textwidth}cccc@{}}
\noalign{\hrule height 1.2pt\vskip 3pt}
\multicolumn{1}{>{\centering\arraybackslash}m{0.14\textwidth}|}{Paradigm} & \multicolumn{1}{>{\centering\arraybackslash}m{0.20\textwidth}}{\shortstack{Representative\\methods / combinations}} & \multicolumn{1}{>{\centering\arraybackslash}m{0.22\textwidth}}{Strategy} & \shortstack{Low\\cost} & \shortstack{Natural\\motion} & \shortstack{Complex-shape\\generation} & \shortstack{Fine-grained\\control} \\
\hline
LLM + specialized motion generator & Motion-Agent~\cite{wu2024motionagent}; motion models such as MoMask~\cite{guo2023momask} and UniMate~\cite{mou2026unimate} & Decomposes complex instructions into motion segments synthesized by specialized models & \tabgood & \tabgood & \tabbad & \tabbad \\
LLM + video generation and motion recovery & Seedance 2.0~\cite{seedanceteam2026seedance2} and Kling-Omni~\cite{klingteam2025klingomni} generate video; DuoMo~\cite{wang2026duomo} and GENMO~\cite{li2025genmo} recover human motion & Plans video motion, recovers part movement from the generated video, and retargets it to a 3D asset & \tabbad & \tabgood & \tabmid & \tabbad \\
Direct LLM motion generation & Claude Opus 5.5, GPT-6; GenRA~\cite{yaroshevych2026genra} & Uses spatiotemporal reasoning to generate motion programs or joint keyframes & \tabmid & \tabmid & \tabgood & \tabgood \\
\noalign{\hrule height 1.2pt}
\end{tabular}
\end{table*}

Recent motion generation models~\cite{guo2023momask,rempe2026kimodo,wang2026motionbricks,mou2026unimate} can synthesize natural and fluent motion from text, yet their controllability remains limited. Given a long-horizon instruction such as ``raise the left wing to 60 degrees within two seconds, flap three times, then glide and land,'' they rarely complete every phase in order or meet the specified timing, angle, and repetition requirements. Moreover, they are typically trained on limited datasets, leaving generalization and complex-instruction handling limited, whereas real 3D assets range from multipeds and axial bodies to winged creatures, mixed-limb characters, and mechanical structures. Existing motion benchmarks share these assumptions, covering conventional skeletons with relatively simple instructions.

Large language models (LLMs) offer a way to handle such complex instructions, and current systems use them in three ways (Table~\ref{tab:intro_paradigms}). A common practice is to let the LLM split the instruction into segments for a specialized motion generator; the result looks natural and is cheap to produce, but remains bound to the generator's skeleton and coarse semantics. Another route has the LLM plan a video, from which 3D motion is recovered and retargeted. Video models bring strong visual priors, but generation is expensive, recovery models are developed mainly for humans, and precise timing and amplitude are often lost along the way. The LLM can also write motion itself through spatiotemporal reasoning, which adapts to arbitrary structures and allows fine-grained control; however, frame-by-frame joint outputs consume many tokens and lack kinematic constraints.

To compare these paradigms under diverse morphologies and complex instructions, we introduce \textit{DivDynBench}, which evaluates rigging and motion generation across six morphology categories: humanoid, multiped, axial, winged, mixed-limb, and mechanical. Its text instructions and numeric constraints require multi-phase part coordination and explicit timing, angle, repetition, and trajectory control. The evaluation separates textual semantic compliance from program-verified numeric accuracy, and further assesses phase completeness, morphology fit, and animation quality. On this benchmark, we systematically analyze the control capabilities, cross-morphology adaptability, and generation costs of the three paradigms.

The comparison shows that direct agent-driven generation holds the most promise for cross-morphology adaptability and fine-grained control, but outputting every joint coordinate frame by frame is costly and difficult, and the lack of bone-length and contact constraints makes the motion prone to interpenetration and hard to control. We therefore propose an Agentic Motion Framework. The agent first divides the instruction into motion phases and timing, and then selects key parts and end-effectors and plans their trajectories and targets in each phase. Next, the executor obtains full-body poses through kinematic solving and corrects interpenetration and ground contact using the deformed mesh. Finally, the exported animation is validated. Because the agent writes only a small set of semantic parameters, this pipeline reduces generation overhead and improves rigging reliability and motion-control accuracy.

Our contributions are as follows:
\begin{itemize}
    \item We analyze complex instruction-driven motion generation for diverse 3D assets, a task that requires multi-part coordination, multi-phase execution, and numeric control. And we summarize three common LLM-assisted paradigms.
    \item We propose \textit{DivDynBench}, a benchmark covering six morphology categories. It evaluates rigging and complex motion generation, using visual scoring and program verification to measure instruction compliance, phase completeness, morphology fit, and animation quality.
    \item We compare the three paradigms on this benchmark in terms of control, cross-morphology adaptability, and generation cost. Based on this, we design an Agentic Motion Framework that improves rigging and motion generation.
\end{itemize}

\bluelight{Relevant methods for dynamic generation and fine-tuning will be integrated into \textbf{HyForge}.}

\section{Related Works}

\subsection{3D and 4D Generation}
3D and 4D content generation has become an active research area. Early zero-shot training methods~\cite{poole2022dreamfusion,wang2023prolificdreamer,zeng2023ipdreamer,tang2026we} optimize individual assets with pretrained 2D diffusion priors, improving fidelity, diversity, and image-guided appearance control. Image-conditioned feed-forward reconstruction and editing methods~\cite{hong2024lrm,tang2024lgm,zhu2026vggtedit} greatly reduce per-asset generation time. Hunyuan3D 1.0~\cite{yang2024hunyuan3d} combines multiview diffusion with feed-forward reconstruction for high-quality asset creation. Subsequent 3D generative models~\cite{zhao2025hunyuan3d,xiang2024trellis} improve geometric detail and appearance, while approaches using 3D latent diffusion and sparse representations~\cite{wu2024direct3d,wu2025direct3ds2} advance high-resolution shape generation.

For dynamic content, 4D reconstruction methods~\cite{xu20244k4d,yang2025widerange4d} recover scenes from multiview videos, achieving high-fidelity, real-time novel-view synthesis and robust reconstruction under wide-range movements and scene changes. Diffusion-based generative methods~\cite{bahmani2023fourdfy,ren2023dreamgaussian4d} synthesize dynamic assets, L4GM~\cite{ren2024l4gm} enables feed-forward reconstruction from monocular video, and Animate3D~\cite{jiang2024animate3d} animates existing assets with multiview video diffusion. These advances improve visual quality and efficiency, but visual realism alone does not ensure execution of complex instructions involving coordinated parts, ordered phases, and precise motion constraints.

\subsection{Agentic Generation}
To handle more complex and precise 3D and 4D generation requests, recent methods introduce language-model agents for planning, tool use, and iterative refinement. Holodeck~\cite{yang2024holodeck} translates language into asset selection and spatial constraints for environment construction, and SceneCraft~\cite{hu2024scenecraft} generates Blender code from scene graphs and refines rendered scenes through visual feedback. Agentic 4D scene composition methods~\cite{xu2024comp4d,zeng2024trans4d} decompose multi-object scenes into trajectory-planned components and model geometry-aware temporal transitions between them. Beyond single-scene construction, world-model unified methods~\cite{guo2026worldclaw,wu2026code2games,gamefactory3arepo,zeng2026worldmodels,zeng2026openworldlib} combine agent planning, generation, and feedback to construct explorable 3D worlds and playable gaming worlds, argue for moving beyond injecting world knowledge into isolated tasks toward a unified framework, and provide standardized codebases and definitions for such advanced world models. Recent coding models such as GPT-6 Astra and Claude Opus 5.5 also demonstrate strong interactive 3D scene-building capabilities in reported AA-3D and professional-environment evaluations~\cite{alphasignalAA3D,gao2026engiworld,gao2026spatialworld}. However, constructing visually rich dynamic scenes remains distinct from generating precise, instruction-following motion for avatars with diverse articulated structures.

\subsection{Avatar Motion Generation}
Motion generation methods explore diffusion-based modeling~\cite{tevet2023mdm} and discrete motion representations~\cite{zhang2023t2mgpt,guo2023momask}. Text-motion datasets~\cite{guo2022humanml3d,yang2024animalml3d} support language-conditioned human and animal motion, while methods~\cite{lee2025dragon,wang2026bimotion,mou2026unimate} extend generation across diverse skeletons. Kimodo~\cite{rempe2026kimodo} advances constrained human motion generation, and MotionBricks~\cite{wang2026motionbricks} combines a modular latent generative model with smart primitives for real-time locomotion and interaction control. Video-to-motion recovery methods~\cite{wang2026duomo,li2025genmo,tao2026skelmo} instead recover 3D motion from video for retargeting.

Animating a given mesh also requires rigging and motion transfer. Automatic rigging methods~\cite{xu2020rignet,zhang2025unirig,liu2025riganything} infer skeletons and skinning, while Puppeteer~\cite{song2025puppeteer} connects rigging with video-guided animation. Neural Kinematic Networks~\cite{villegas2018nkn} transfer motion across body proportions, and Skeleton-Aware Networks~\cite{aberman2020skeletonaware} support retargeting between different but homeomorphic skeletons. Agent-driven animators~\cite{yaroshevych2026genra,zou2026artimo} require pre-rigged or URDF-articulated assets and do not handle long multi-joint limbs. Despite these advances in control and skeletal coverage, reliably executing long-horizon, multi-part instructions on nonstandard skeletons remains challenging. Recent benchmarks~\cite{yang2026t2mbench,lin2025mbench} examine instruction following and motion quality, while \textit{DivDynBench} evaluates rigging and complex text/numeric motion across diverse morphologies.

\section{\textit{DivDynBench}}

\begin{figure*}[t]
\centering
\includegraphics[width=\textwidth]{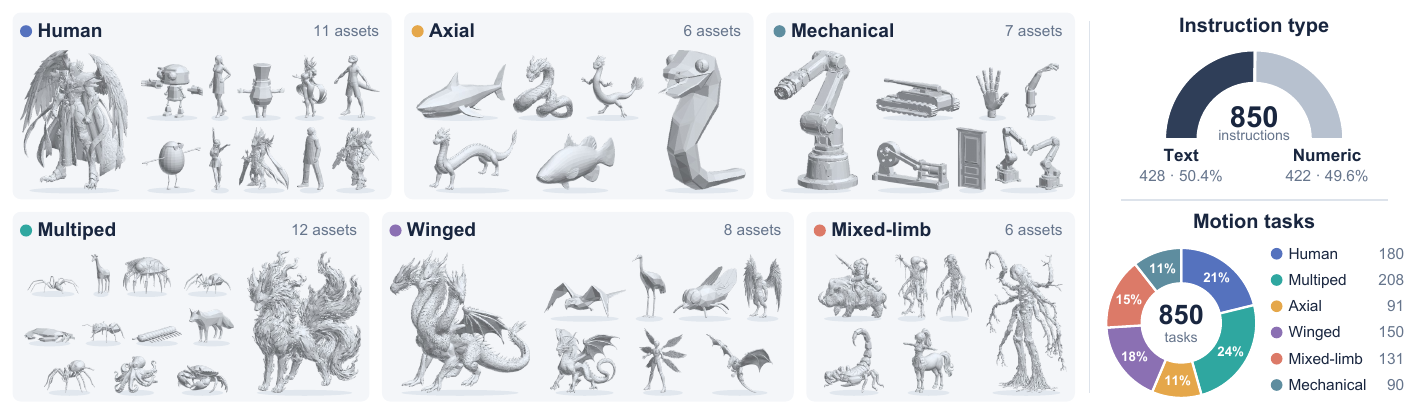}
\caption{Overview of \textit{DivDynBench}. The left panel displays white-mesh renderings of all 50 assets. The center panel compares the 428 text and 422 numeric instructions, while the right panel shows how the 850 motion tasks are distributed across six asset categories.}
\label{fig:bench_assets}
\end{figure*}

\begin{table}[t]
\centering
\scriptsize
\setlength{\tabcolsep}{2pt}
\renewcommand{\arraystretch}{1.04}
\caption{Comparison of motion benchmarks by test size, subjects, and instruction type.}
\label{tab:bench_comparison}
\begin{tabular}{@{}>{\raggedright\arraybackslash}p{0.23\columnwidth}>{\centering\arraybackslash}p{0.10\columnwidth}>{\raggedright\arraybackslash}p{0.32\columnwidth}>{\raggedright\arraybackslash}p{0.29\columnwidth}@{}}
\hline
Benchmark & Cases & Subjects & Instruction type \\
\hline
AnimalML3D~\cite{yang2024animalml3d} & 318 & Animal (SMAL) & Short descriptive \\
M-Bench~\cite{lin2025mbench} & 450 & Human & Long descriptive \\
T2MBench~\cite{yang2026t2mbench} & 1,025 & Human & Multi-stage descriptive + numeric pose \\
ArtiMo~\cite{zou2026artimo} & 225 & Articulated objects & Phase-ordered joint commands \\
\hline
\textbf{\textit{DivDynBench}} & \textbf{850} & \textbf{Humanoid, multiped, axial, winged, mixed-limb, mechanical} & \textbf{Multi-phase descriptive + timed numeric} \\
\hline
\end{tabular}
\end{table}

\begin{figure}[t]
\centering
\includegraphics[width=0.92\columnwidth]{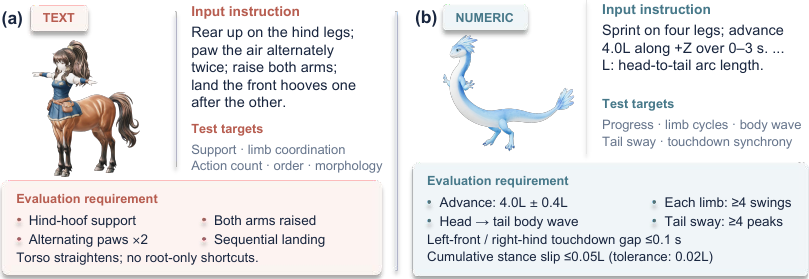}
\caption{Representative text (a) and numeric (b) tasks in \textit{DivDynBench}. Each example shows the asset and an abbreviated instruction, together with test targets and selected evaluation requirements withheld from the generator. The numeric example requires coordinated limb motion, a head-to-tail body wave, and synchronized touchdown.}
\label{fig:bench_case}
\end{figure}

\textit{DivDynBench} follows a common 3D asset import workflow: its inputs are readable GLB assets, rather than ground-truth motion clips on a standardized human skeleton. We define a rigging task starting from a raw mesh and an instruction-driven motion task using either a predicted rig or a provided reference rig.

\subsection{Asset Collection and Morphology Coverage}
To reflect practical user workflows, we collect outputs from 3D generative models, basic shapes downloaded directly from the internet, and assets from engine libraries. These cover humanoids, multipeds, axial bodies, winged creatures, mixed-limb characters, and articulated mechanical objects. Each asset is organized as a self-contained GLB with a manually designed skeleton and skinning weights as rigging ground truth. The 50 assets and their associated 850 motion tasks are shown in Fig.~\ref{fig:bench_assets}.

Table~\ref{tab:bench_comparison} compares motion-generation case counts, morphology coverage, and task complexity. \textit{DivDynBench} provides a rich collection of motion test cases with the broadest skeletal-morphology coverage among the compared benchmarks. Its challenging instructions demand coordinated multi-part motion, ordered multi-phase execution, and precise adherence to temporal and numerical constraints.

\subsection{Task Construction and Instruction Types}
Each task pairs an asset with a motion instruction, itemized scoring criteria, and structured rules. Text tasks emphasize part coordination, action order, and phases. Numeric tasks additionally specify timing, angles, displacements, repetition counts, or tolerances, with pre-reviewed check specifications.

The text task in Fig.~\ref{fig:bench_case}(a) uses mainly qualitative descriptions rather than specifying every motion parameter precisely. Its combined requirements for multi-part coordination and multi-stage execution test whether a model can translate compositional semantics into complete actions. Numeric tasks instead test precise adherence to specified angles, displacements, and timing, rather than merely generating semantically similar motion. Fig.~\ref{fig:bench_case}(b) illustrates requirements on displacement, rhythm, and synchrony.

\subsection{Evaluation Metrics}
We evaluate structural inference and motion control through rigging and motion generation metrics. Evaluation combines programmatic measurements of geometry, motion, and numeric constraints with rubric-based visual scoring by a multimodal language model.

\paragraph{Rigging.}
Rigging evaluation metrics include Source Fidelity, J2J, mIoU, Joint validity, Edge $p_{99}$, and Rigid RMS. Source Fidelity measures the proportion of valid rigs that keep the source mesh, while J2J and mIoU measure joint-position error and skinning segmentation agreement against manually designed ground truth. Joint validity measures the fraction of joints inside or near the mesh surface. Edge $p_{99}$ and Rigid RMS measure edge-length changes under standard poses and non-rigid distortion of mechanical parts, respectively. All are computed programmatically.

\paragraph{Motion.}
Motion evaluation metrics include Semantic, Accuracy, Completeness, Morphology, Quality, Aesthetics, and Overall. Semantic assesses text instruction following against the scoring rubric, while Accuracy measures numeric constraint satisfaction programmatically. Completeness assesses phase order and completion through visual scoring for text tasks and programmatic checks for numeric tasks. Morphology and Aesthetics use visual scoring, with additional programmatic constraints on Morphology. Quality is computed from penetration, foot sliding, deformation, and smoothness tests. Scores are out of 100, and Overall is a weighted score computed per case. \textbf{Metric definitions and scoring details are provided in the supplementary material.}

\section{Agentic Motion Framework}

\begin{figure*}[t]
\centering
\includegraphics[width=0.92\textwidth]{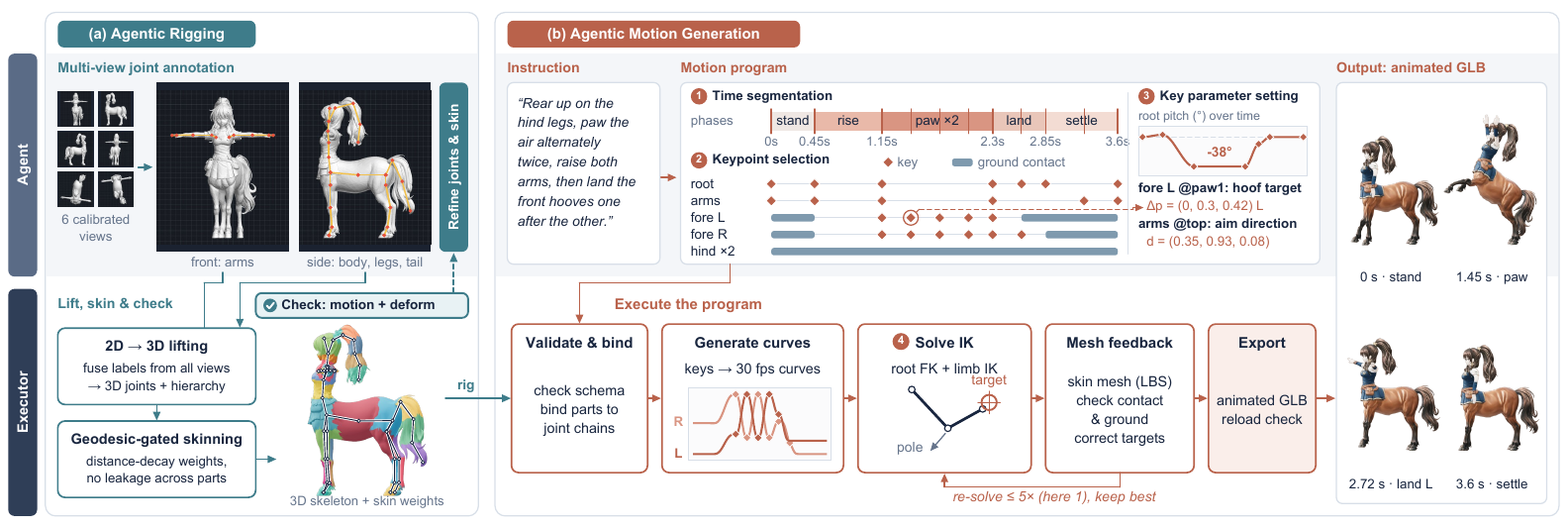}
\caption{Agentic Motion Framework. The agent writes skeleton configurations and parametric motion programs; the executor reconstructs the skeleton, computes skinning weights, solves kinematics, and exports animations. A provided rig bypasses the rigging stage.}
\label{fig:agent_overview}
\end{figure*}

\begin{table*}[t]
\centering
\small
\caption{Rigging of raw 3D assets (49 of 50 assets; one of two similarly shaped spiders is excluded). Source Fidelity is category-macro-averaged; mIoU and joint validity assign zero to failures. Rigid RMS applies only to mechanical assets.}
\label{tab:rigging_main}
\begin{tabular}{lcccccc}
\hline
Method & Source Fidelity (\%)$\uparrow$ & J2J$\downarrow$ & mIoU$\uparrow$ & Joint validity$\uparrow$ & Edge $p_{99}\downarrow$ & Rigid RMS$\downarrow$ \\
\hline
Puppeteer~\cite{song2025puppeteer} & 98.48 & 0.0707 & 0.2738 & 0.8631 & 0.5951 & 0.0509 \\
SkinTokens~\cite{zhang2026skintokens} & 15.43 & 0.0534 & 0.0632 & 0.1504 & 0.4401 & \textbf{0.0168} \\
UniRig~\cite{zhang2025unirig} & 13.92 & 0.0696 & 0.0469 & 0.1316 & 0.3852 & 0.0490 \\
RigAnything~\cite{liu2025riganything} & 15.43 & 0.0690 & 0.0579 & 0.1369 & 0.4078 & 0.0295 \\
GPT6.1-sol & 67.82 & 0.0749 & 0.2316 & 0.4048 & 0.2529 & 0.0569 \\
GPT6.1-sol + Ours & \textbf{100.00} & \textbf{0.0389} & \textbf{0.4174} & \textbf{0.9718} & 0.2534 & 0.0265 \\
Claude-ops5.5 & 66.29 & 0.0565 & 0.2610 & 0.5920 & \textbf{0.2484} & 0.0415 \\
Claude-ops5.5 + Ours & \textbf{100.00} & 0.0410 & 0.3846 & 0.9209 & 0.2493 & 0.0209 \\
\hline
\end{tabular}
\end{table*}

\begin{table*}[t]
\centering
\scriptsize
\caption{Motion generation on reference rigs across six asset categories (428 text and 422 numeric cases; per-case mean, out of 100). Semantic measures text compliance, while Accuracy measures program-verified numeric accuracy. Cost is the per-case generation cost in USD. Bold denotes the best value within each paradigm.}
\label{tab:motion_general}
\begin{tabular}{lcccccccc}
\hline
Method & Semantic$\uparrow$ & Accuracy$\uparrow$ & Completeness$\uparrow$ & Morphology$\uparrow$ & Quality$\uparrow$ & Aesthetics$\uparrow$ & Overall$\uparrow$ & Cost (USD)$\downarrow$ \\
\hline
\multicolumn{9}{c}{\textit{LLM + specialized motion generator}} \\
\hline
GPT6.1-sol + UniMate~\cite{mou2026unimate} & 0.46 & \textbf{7.69} & 3.69 & 32.38 & \textbf{44.43} & \textbf{40.56} & \textbf{19.95} & \textbf{0.030} \\
Claude-ops5.5 + UniMate~\cite{mou2026unimate} & \textbf{2.77} & 7.36 & \textbf{4.41} & \textbf{33.82} & 39.32 & 37.07 & 19.29 & 0.045 \\
\hline
\multicolumn{9}{c}{\textit{LLM + video generation and motion recovery}} \\
\hline
Seedance 2.5~\cite{seedteam2026seedance25} + GPT6.1-sol & 10.56 & 9.05 & 8.90 & 40.18 & 31.96 & \textbf{49.12} & 22.55 & \textbf{0.389} \\
Seedance 2.5~\cite{seedteam2026seedance25} + Claude-ops5.5 & \textbf{13.55} & \textbf{11.00} & \textbf{11.27} & \textbf{49.52} & \textbf{44.70} & 42.16 & \textbf{27.14} & 0.418 \\
\hline
\multicolumn{9}{c}{\textit{Direct LLM motion generation}} \\
\hline
GPT6-astra & 17.76 & 18.46 & 17.21 & 46.01 & \textbf{57.76} & 44.03 & 32.64 & 0.050 \\
GPT6.1-sol & 20.15 & 17.27 & 18.25 & 45.96 & 46.53 & 37.86 & 30.19 & 0.024 \\
Claude-ops5.5 & 45.68 & 47.03 & 46.08 & 51.25 & 49.06 & \textbf{52.52} & 48.19 & 0.065 \\
GPT6.1-sol + Ours & 23.25 & 26.13 & 24.53 & 46.51 & 52.57 & 49.58 & 35.99 & \textbf{0.0087} \\
Claude-ops5.5 + Ours & \textbf{45.84} & \textbf{50.72} & \textbf{47.89} & \textbf{53.24} & 52.35 & 51.44 & \textbf{50.07} & 0.047 \\
\hline
\end{tabular}
\end{table*}

\begin{table*}[t]
\centering
\scriptsize
\caption{Motion generation on humanoid assets (90 text and 90 numeric cases; per-case mean, out of 100). Bold denotes the best value within each paradigm.}
\label{tab:motion_human}
\begin{tabular}{lccccccc}
\hline
Method & Semantic$\uparrow$ & Accuracy$\uparrow$ & Completeness$\uparrow$ & Morphology$\uparrow$ & Quality$\uparrow$ & Aesthetics$\uparrow$ & Overall$\uparrow$ \\
\hline
\multicolumn{8}{c}{\textit{Specialized motion generation}} \\
\hline
GPT6.1-sol + MoMask~\cite{guo2023momask} & 7.00 & 4.08 & 5.42 & 43.17 & 30.24 & 47.56 & 20.23 \\
Claude-ops5.5 + MoMask~\cite{guo2023momask} & 10.83 & 7.33 & 8.42 & 43.53 & 46.64 & 44.44 & 25.17 \\
Claude-ops5.5 + MotionBricks~\cite{wang2026motionbricks} & \textbf{52.33} & \textbf{38.05} & \textbf{45.82} & \textbf{55.77} & \textbf{51.96} & \textbf{53.48} & \textbf{49.09} \\
\hline
\multicolumn{8}{c}{\textit{LLM + video generation and motion recovery}} \\
\hline
Seedance 2.5~\cite{seedteam2026seedance25} + DuoMo~\cite{wang2026duomo} & \textbf{7.68} & 10.59 & \textbf{8.73} & 18.54 & \textbf{39.23} & 32.58 & \textbf{18.80} \\
Seedance 2.5~\cite{seedteam2026seedance25} + GenMo~\cite{li2025genmo} & 4.94 & \textbf{12.15} & 8.15 & \textbf{25.97} & 29.01 & \textbf{37.96} & 18.05 \\
\hline
\multicolumn{8}{c}{\textit{Direct LLM motion generation}} \\
\hline
Claude-ops5.5 + Ours & \textbf{54.67} & \textbf{58.03} & \textbf{58.00} & \textbf{59.37} & \textbf{58.88} & \textbf{52.69} & \textbf{57.27} \\
\hline
\end{tabular}
\end{table*}

Generating joint coordinates for diverse skeletons is costly and difficult to control, often causing interpenetration and foot sliding. We instead let a vision-language agent write compact skeleton configurations and parametric motion programs, which a deterministic executor turns into rigs or animations, as shown in Fig.~\ref{fig:agent_overview}. Given a mesh $M=(V,F)$ and instruction $I$, their interfaces are
\begin{equation}
\begin{aligned}
S &= \mathcal A_{\mathrm{rig}}(\mathcal O_M), &
R &= \mathcal X_{\mathrm{rig}}(M,S),\\
P &= \mathcal A_{\mathrm{mot}}(I,\mathcal O_R), &
Y &= \mathcal X_{\mathrm{mot}}(M,R,P).
\end{aligned}
\label{eq:agent_interfaces}
\end{equation}
Here, $\mathcal A$ and $\mathcal X$ denote the agent and executor. The agent observes calibrated asset views $\mathcal O_M$, or views with rig information $\mathcal O_R$, to write rigging configuration $S$ and parametric motion program $P$. The executor produces rig $R$ and animation $Y$. A provided rig sets $R=R_{\mathrm{in}}$ and bypasses rigging.

\subsection{Agentic Rigging}
\paragraph{Skeleton construction.}
In the rigging stage of Fig.~\ref{fig:agent_overview}(a), the agent identifies parts in calibrated views, annotates joints, and connects them into bone chains. The executor reconstructs 3D joint positions from these annotations and builds the specified hierarchy. Part-to-joint mappings associate instruction terms with the asset's actual structure, without assuming a fixed human skeleton.

\paragraph{Skinning and validation.}
The executor computes distance-based skinning weights and smooths them within geodesically gated neighborhoods to limit influence across separate appendages. The agent can restrict influences by part and assign rigid components to individual bones. The rig is then reloaded and checked with probe deformations to guide skeleton and skinning refinement. \textbf{Detailed formulations are provided in the supplementary material.}

\subsection{Agentic Motion Generation}
Fig.~\ref{fig:agent_overview}(b) outlines how the agent plans phase timing, selects the parts to move, and specifies their trajectories and targets. These decisions form a compact program:
\begin{equation}
\begin{aligned}
P&=\left(T,f,\mathcal H,\mathcal K,\mathcal R_0,\{P_p\}_{p\in\mathcal Q},\xi\right),\\
\mathcal R_0&=(j_0,o_0,\kappa_0),\quad
P_p=\left(\Gamma_p,o_p,\kappa_p,\mathcal W_p\right).
\end{aligned}
\label{eq:agent_motion_program}
\end{equation}
Here, $T,f$ specify duration and frame rate, $\mathcal H,\mathcal K$ specify phase windows and key events, and $\xi$ configures refinement. Root control $\mathcal R_0$ applies operator $o_0$ with parameters $\kappa_0$ to root $j_0$. For each part $p\in\mathcal Q$, $P_p$ specifies a joint chain $\Gamma_p$, operator $o_p$, motion parameters $\kappa_p$, and contact windows $\mathcal W_p$.

\paragraph{Phase timing.}
The agent sets phase boundaries and repetition counts to organize action order and rhythm. A control curve within phase $[a_h,b_h]$ uses local progress. For an endpoint transition,
\begin{equation}
\begin{aligned}
\phi_h(t)&=\operatorname{clip}\!\left(\frac{t-a_h}{b_h-a_h},0,1\right),\\
\mathbf c_h(t)&=\mathbf c_h^0+(\mathbf c_h^1-\mathbf c_h^0)\,e_h(\phi_h(t)).
\end{aligned}
\label{eq:agent_phase_progress}
\end{equation}
Here, $\mathbf c_h^0,\mathbf c_h^1$ are endpoint values and $e_h$ controls easing. Phase duration and easing determine transition speed, while count or frequency parameters control repeated actions.

\paragraph{Part trajectories and targets.}
Using the rig's part mappings, the agent assigns motion to the relevant chains and end-effectors. Root parameters $\kappa_0$ describe global displacement and orientation, while part parameters $\kappa_p$ specify target positions, angles, or trajectories over the planned phases. Contact windows identify when an end-effector should remain anchored. This separates global motion from local coordination and allows edits to individual phases or parts.

\paragraph{Chain solving.}
Existing agent-driven methods~\cite{yaroshevych2026genra,zou2026artimo} depend on pre-built rigged USD or URDF assets and set joint values one by one, which scales poorly to long limbs such as tails and tentacles. Our end-effector operator instead bends a whole chain by one scalar distributed through per-joint weights, so the agent shapes the motion of an entire chain of any length by adjusting only its end-effector trajectory through timed keys, contacts, and gait cycles, which supports complex motion on diverse chains. Each joint is placed at its parent's position plus its rotated bone vector, so bone lengths never change. For an unreachable target, the bend stays within joint limits, the chain reaches the closest attainable distance while still pointing at the target, and the residual is passed to refinement. \textbf{Due to space limits, detailed formulations are provided in the supplementary material.}

\paragraph{Execution and validation.}
The executor is a parameterized Python program, \textit{independent of our evaluation pipeline}, invoked directly by the agent as a tool. It solves joint poses with inverse and forward kinematics, deforms the mesh through skinning, and revises the input parameters based on mesh feedback to reduce penetration, contact errors, and end-effector deviations.

\section{Experiments}

\begin{figure*}[t]
\centering
\includegraphics[width=0.92\textwidth]{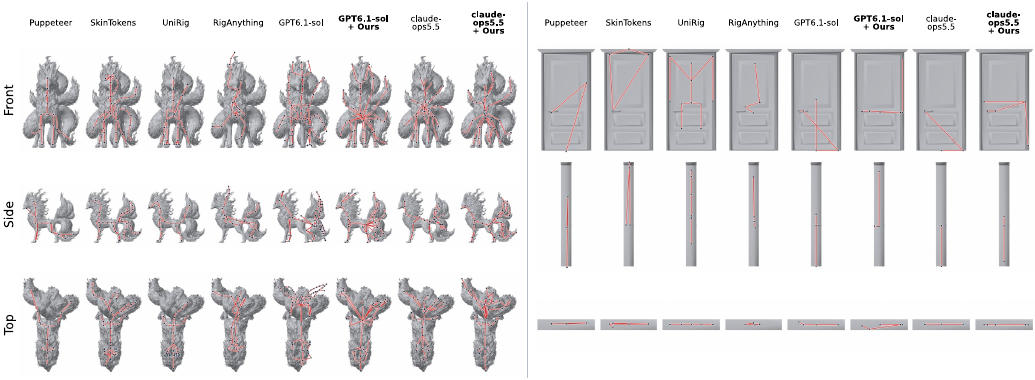}
\caption{Comparison of skeletons, skinning weights, and probe poses for the same 3D asset.}
\label{fig:rigging_compare}
\end{figure*}

\begin{figure*}[t]
\centering
\includegraphics[width=0.92\textwidth]{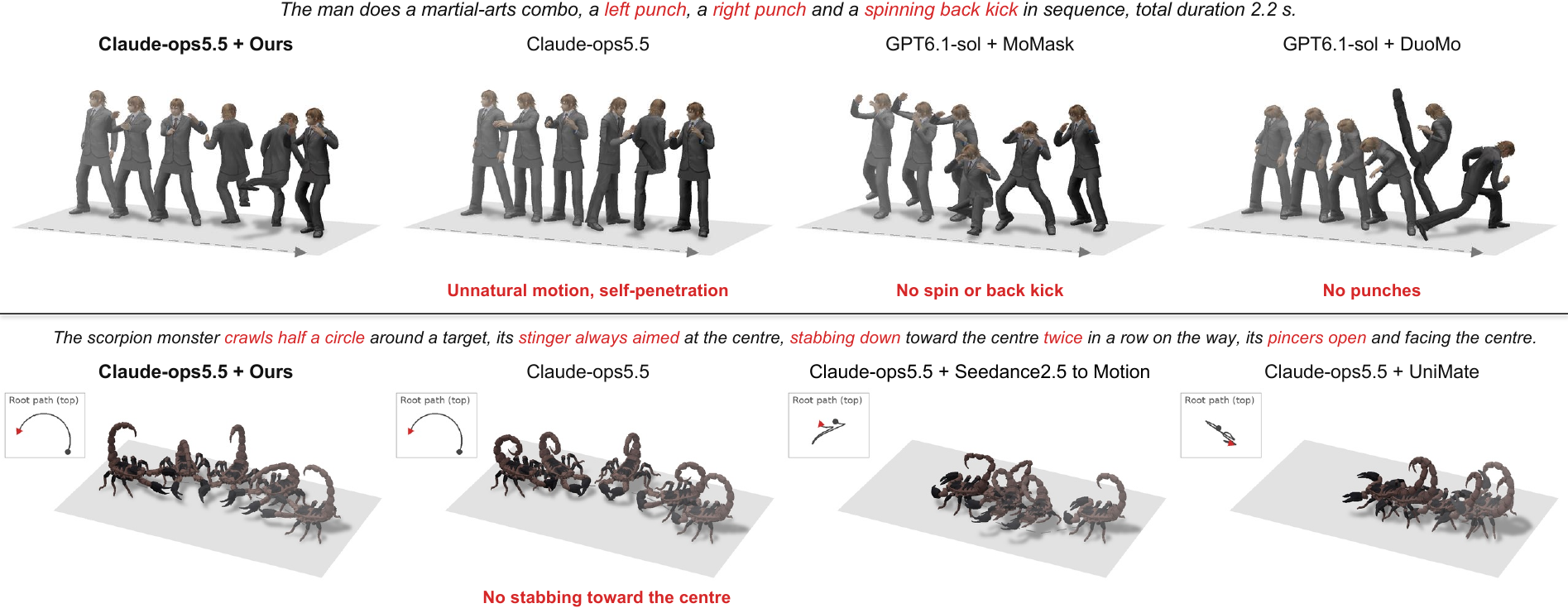}
\caption{Key-pose comparison for the same 3D asset and motion instruction.}
\label{fig:motion_compare}
\end{figure*}

\begin{figure}[t]
\centering
\includegraphics[width=\columnwidth]{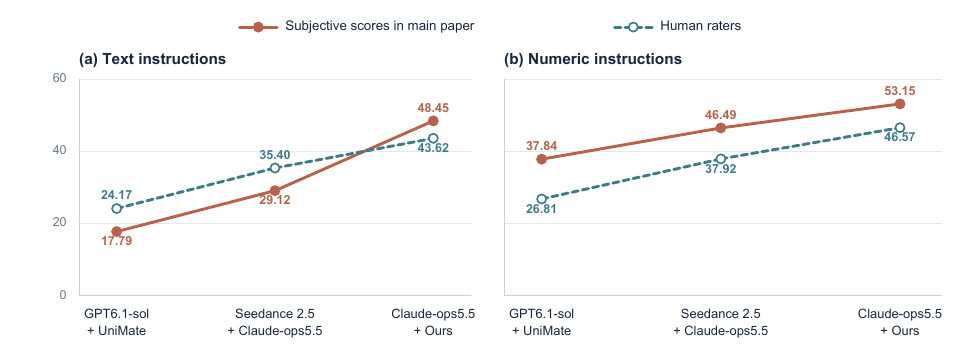}
\caption{Comparison between user-study ratings and GPT6.1-sol scores used in this paper.}
\label{fig:user_study}
\vspace{-4mm}
\end{figure}

We evaluate rigging and motion generation separately. Motion metrics such as morphology fit, penetration, foot sliding, and deformation are computed on the skinned mesh, so a predicted rig would entangle rigging failures with motion-planning failures. We therefore assess rigging from raw meshes and evaluate motion on a provided reference rig, so that motion scores reflect instruction following and kinematic control alone. End-to-end results that animate predicted rigs are reported in the supplementary material.

\subsection{Compared Methods and Scoring}
For rigging, we compare specialized rig methods~\cite{song2025puppeteer,zhang2026skintokens,zhang2025unirig,liu2025riganything} with GPT6.1-sol and Claude-ops5.5 writing skeleton configurations directly. For both rigging and motion, the same base agent with and without the Agentic Motion Framework keeps \textit{the same setup and a comparable number of turns} as its skill-free counterpart, so the pair is \textit{fair}. For motion, we group baselines by the three paradigms in Table~\ref{tab:intro_paradigms}: an LLM driving a specialized generator~\cite{mou2026unimate}, Seedance 2.5 generating a motion video that is then recovered via a tracking model with LLM-based adjustment, and LLMs (GPT6.1-sol, Claude-ops5.5, GPT6-astra) generating motion directly. Comparing paradigms under the same instructions shows which route can actually satisfy multi-phase and numeric requirements on diverse morphologies. Humanoid-only methods~\cite{guo2023momask,wang2026motionbricks,wang2026duomo,li2025genmo} are compared separately on the humanoid subset.

Rigging metrics are computed programmatically against the manually designed ground truth. For motion, every result is exported as an animated GLB on the reference rig, rendered from fixed views, and scored with the rubric-based and programmatic metrics of \textit{DivDynBench} on a 0-100 scale. Semantic and Accuracy average text and numeric cases, respectively, while the other dimensions and the per-case weighted Overall average all applicable cases, with invalid cases counted as zero. Generation receives only the source asset, the instruction, and, for motion, the reference rig, and all evaluation materials are fixed before generation.

\subsection{Rigging Evaluation}
Table~\ref{tab:rigging_main} shows that equipping both agents with the Agentic Motion Framework makes them reliable at rigging raw assets, and the gains in Source Fidelity, J2J, mIoU, and joint validity are clearly visible. SkinTokens, UniRig, and RigAnything all process the input mesh, so their outputs differ from the source mesh that the rig must drive, which leads to their low Source Fidelity. Puppeteer keeps the mesh but still trails on J2J, mIoU, and joint validity. Edge $p_{99}$ is essentially unchanged and Rigid RMS improves toward the best specialized value, so the gains do not come at the cost of larger deformation.

Fig.~\ref{fig:rigging_compare} shows the skeletons predicted for a nine-tailed fox and a door in front, side, and top views. On the fox, the specialized methods produce sparse skeletons concentrated on the trunk and legs and leave most tails unrigged. GPT6.1-sol alone places several chains outside the mesh, most visibly beyond the tails in the top view, and Claude-ops5.5 alone again misses most tails. With the Agentic Motion Framework, both agents grow a separate chain into each tail while keeping the legs, head, and spine inside the body, and the skeleton stays consistent across all three views. On the door, the baselines draw diagonal bones across the panel or, in the case of UniRig, a humanoid-like tree, whereas with the framework the agents place the main bone along the hinge edge and a short bone at the handle, which matches the door's rigid articulation and explains the lower Rigid RMS. These gains come from multimodal observation. The agent labels each joint in the view where the part is clearest, such as the top view for separating tails and the side view for the legs, and the executor fuses these labels into a 3D skeleton that follows the asset's actual morphology rather than a template.

\subsection{Motion Generation Evaluation}
Table~\ref{tab:motion_general} shows that the three paradigms differ mainly in controllability. Specialized generators and video recovery produce plausible poses but rarely satisfy the ordered phases and numeric targets, since text segments or generated videos cannot impose a required angle. Direct LLM generation can instead design motion according to the specified angles, but its results vary strongly across agents. The Agentic Motion Framework makes such angle-based design stable by attaching root orientation, joint rotations, and end-effector targets to key events in the program, and the table shows clear gains for both agents, especially in Accuracy, Quality, and Overall. The comparison also shows that direct LLM generation is very economical, and using our framework lowers its cost further. The supplementary material compares against a variant without chain solving and against ArtiMo~\cite{zou2026artimo}, whose articulation model is more restricted.

To examine humanoid-specific motion generators, we compare MoMask, MotionBricks, and video-based recovery on the humanoid subset in Table~\ref{tab:motion_human}. MoMask and the recovery methods still score below 26 overall even on this familiar morphology, showing that segment-level prompting and recovered video motion do not enforce the full instruction or its angles. MotionBricks deserves separate attention because it supports trajectory guidance. When a Claude-ops5.5 agent supplies structured IK targets and trajectories, it reaches 49.09 overall and leads all seven columns among the specialized generators. This large gain over MoMask shows that letting an agent plan trajectories while a specialized module solves them is an effective division of labor, although MotionBricks itself remains limited to humanoid skeletons. Claude-ops5.5 + Ours follows the same division of labor, but supports arbitrary skeletons and controls rhythm through explicit phases and key events, raising the score further to 57.27 on the same subset. This supports separating parametric program authoring from deterministic execution.

Fig.~\ref{fig:motion_compare} shows key poses for two instructions. The humanoid must throw a left punch, a right punch, and a spinning back kick in sequence within 2.2~s. Claude-ops5.5 + Ours raises its guard, throws both punches, turns its back during the spin, extends the leg backward for the kick, and recovers. Claude-ops5.5 without the framework produces unnatural poses with self-penetration. GPT6.1-sol + MoMask stays in guard and crouch poses with neither spin nor back kick, and GPT6.1-sol + DuoMo performs a high kick and a lunge but no punches. The two humanoid-specific pipelines thus each drop a phase of the sequence. The scorpion must crawl half a circle around a target and stab toward the centre twice on the way. Claude-ops5.5 + Ours follows a half-circle root path and lowers its stinger forward to stab. Claude-ops5.5 without the framework follows a similar path, yet its stinger stays curled upward and never stabs, showing that direct generation can capture the global trajectory but loses the local part motion. The Seedance-based recovery and UniMate move the root only along a short zigzag or hooked segment, and the body stays in place instead of circling the target. In both examples, explicit phase timing, root trajectories, and part targets in the motion program let the agent satisfy the global path and the local actions together.

\subsection{User Study}
To verify the reasonableness of scoring with GPT6.1-sol, we conduct a user study in which 30 volunteers rate each result on Semantic, Completeness, Morphology, and Aesthetics for GPT6.1-sol + UniMate, Seedance 2.5 + Claude-ops5.5, and Claude-ops5.5 + Ours. The subjective score averages all four dimensions for text tasks, and only Morphology and Aesthetics for numeric tasks, whose Accuracy and Completeness are computed programmatically. As shown in Fig.~\ref{fig:user_study}, the volunteer ratings follow the same trend as the LLM scores on both tasks and rank the three methods in the same order, confirming the reasonableness of the original scoring.

\section{Conclusion}

We find that LLMs can already animate 3D assets directly from complex instructions, but not reliably. Rigging and complete multi-phase motion succeed only intermittently, with large variation across agents and morphologies. Existing benchmarks, built around conventional skeletons and simple prompts, do not expose this instability. \textit{DivDynBench} closes that gap across six morphology categories. Comparing three LLM-based paradigms, we find specialized models efficient but skeleton-bound, video generation with motion recovery visually strong yet imprecise in timing and amplitude, and direct LLM-agent generation comparatively the most viable, though still unstable. Our Agentic Motion Framework, which couples agent-authored parametric programs with deterministic rigging and kinematics, markedly improves rigging quality and the controllability and stability of motion generation using the same base agents. We hope the benchmark and framework support the deployment of instruction-driven animation for diverse 3D assets.

\section*{LLM Usage Statement}

Large language models (LLMs) were used as general-purpose writing assistants for this manuscript, performing language polishing and grammar checking through Codex, an LLM-based coding agent. They did not contribute to research ideation, method design, experimental implementation, data analysis, result interpretation, or the drafting of substantive content. All scientific content, claims, and conclusions were conceived and written by the authors, who reviewed every LLM-suggested edit and take full responsibility for the paper. The use of LLM agents as subjects of evaluation in our benchmark and framework is part of the research itself and is described in this paper.

{
    \small
    \bibliographystyle{ieeenat_fullname}
    \bibliography{main}
}

\clearpage
\maketitlesupplementary

\begin{table*}[t]
\centering
\caption{Baseline setup and parameters. All methods use essentially the same setting: the same rigged assets and benchmark cases, and rig-compatible 30\,fps GLB motion as output. Seedance refers to Seedance 2.5 image-to-video followed by CoTracker3 tracking and known-rig fitting. The video-to-motion pipelines use the same agent as the other methods to post-process the generated results. For Ours, the IK regularization and ground weight correspond to $\lambda_{\mathrm{IK}}$ and $\lambda_g$ in the main text. Program checks are checking programs that each agent builds from the input requirements, unrelated to evaluation.}
\label{tab:supp_baseline_setup}
\scriptsize
\setlength{\tabcolsep}{3pt}
\renewcommand{\arraystretch}{1.1}
\newcolumntype{Z}{>{\raggedright\arraybackslash}p{0.155\textwidth}}
\begin{tabular}{@{}>{\raggedright\arraybackslash}p{0.11\textwidth} Z Z Z Z Z@{}}
\toprule
 & \textbf{Agent w/o skills} & \textbf{Ours} & \textbf{UniMate} & \textbf{Seedance + track} & \textbf{MoMask / MotionBricks / DuoMo / GENMO} \\
\midrule
Backbone
 & Same agent, self-written FK scripts
 & Same agent; fills program parameters
 & UniMate + same agent; agent writes prompts + fit
 & Seedance 2.5 I2V + CoTracker3 offline + agent fit
 & Released checkpoints + same agent; MotionBricks adds agent-authored IK recipe; DuoMo/GENMO take Seedance video \\
Input
 & Instruction, rig (self-parsed phases/windows)
 & Instruction, rig (+ numeric spec)
 & Instruction, rig
 & Rig render + instruction prompt
 & Text segments + durations (video for DuoMo/GENMO) \\
Feedback
 & Program checks
 & Executor checks and Geometry checks
 & Geometry checks
 & Program and Geometry checks
 & Program checks \\
Turns (max / mean)
 & 4 / 2.16
 & 4 / 2.38
 & 3 / 1.88
 & 3 / 2.26
 & 5 / 1.64 \\
Key params
 & 30\,fps; jump filter 45$^\circ$
 & IK reg.\ $10^{-3}$; flexion in $(\pm180^\circ)$
 & CFG 3.0; window 60 / overlap 10
 & 400 pts; tracker 512\,px; vis.\ 0.9
 & MoMask $T$=18, cond.\ 4.0; MotionBricks seg.\ 0.55\,s \\
\bottomrule
\end{tabular}
\end{table*}

\begin{table}[t]
\centering
\scriptsize
\setlength{\tabcolsep}{3pt}
\caption{Quality terms for ground-legged assets. Weights change with the environment of the asset.}
\label{tab:supp_quality}
\resizebox{\columnwidth}{!}{%
\begin{tabular}{llccc}
\hline
Term & Measurement & $f$ & $z$ & Weight \\
\hline
Ground penetration & Maximum depth below ground in $L$ & 0.01 & 0.05 & 20 \\
Foot skating & Ratio of sliding contact frames & 0.05 & 0.30 & 20 \\
Floating & Ratio of unsupported airborne frames & 0.02 & 0.20 & 10 \\
Self-penetration & Maximum depth between parts in $L$ & 0.01 & 0.05 & 20 \\
Geometry & Rigid edge change / edge stretch $p_{99}$ & 0.01 / 0.05 & 0.05 / 0.20 & 15 \\
Smoothness & Ratio of acceleration spikes & 0.01 & 0.10 & 15 \\
\hline
\end{tabular}%
}
\end{table}

\begin{table}[t]
\centering
\scriptsize
\caption{User study setup.}
\label{tab:supp_user_study}
\begin{tabular}{@{}>{\raggedright\arraybackslash}p{0.32\columnwidth}>{\raggedright\arraybackslash}p{0.58\columnwidth}@{}}
\hline
Item & Setting \\
\hline
\itshape
Participants & 30 volunteers \\
Compared methods & GPT6.1-sol + UniMate, Seedance 2.5 + Claude-ops5.5, Claude-ops5.5 + Ours \\
Scoring dimensions and scale & Text: Semantic, Completeness, Morphology, Aesthetics (0--100 each); Numeric: Morphology, Aesthetics (0--100 each) \\
Case count and stratification & 150 cases sampled per volunteer from the full case pool \\
Method anonymization and order randomization & The three methods are anonymized as A/B/C, with their presentation order independently randomized for each case \\
Workload per rater & Each case presents all three anonymized outputs together with the scoring rubric, and raters score them by direct comparison \\
Workload and compensation & On average about 4 hours per volunteer to complete all ratings, compensated CNY 200 per person, and no volunteer raised any objection to this payment \\
\hline
\end{tabular}
\end{table}

\begin{table*}[t]
\centering
\scriptsize
\caption{End-to-end motion generation with predicted rigs, by asset category (per-case mean, out of 100). Reference denotes the provided rig used in the main paper; results are shown for the three categories with predicted-rig evaluation.}
\label{tab:supp_e2e}
\begin{tabular}{llccccccc}
\hline
Method & Rig & Semantic$\uparrow$ & Accuracy$\uparrow$ & Completeness$\uparrow$ & Morphology$\uparrow$ & Quality$\uparrow$ & Aesthetics$\uparrow$ & Overall$\uparrow$ \\
\hline
\multicolumn{9}{c}{\textit{Humanoid}} \\
\hline
Claude-ops5.5 + Ours & Reference & 54.67 & 58.03 & 58.00 & 59.37 & 58.88 & 52.69 & 57.27 \\
Claude-ops5.5 + Ours & Predicted & 47.17 & 46.60 & 47.25 & 42.97 & 61.32 & 52.27 & 49.80 \\
\hline
\multicolumn{9}{c}{\textit{Axial}} \\
\hline
Claude-ops5.5 + Ours & Reference & 35.00 & 36.32 & 33.09 & 44.33 & 50.10 & 46.07 & 40.37 \\
Claude-ops5.5 + Ours & Predicted & 24.78 & 27.12 & 25.05 & 36.82 & 50.09 & 49.54 & 34.58 \\
\hline
\multicolumn{9}{c}{\textit{Mixed-limb}} \\
\hline
Claude-ops5.5 + Ours & Reference & 43.69 & 48.39 & 46.67 & 46.73 & 55.65 & 47.32 & 48.33 \\
Claude-ops5.5 + Ours & Predicted & 65.85 & 37.51 & 49.84 & 47.60 & 54.25 & 51.94 & 51.20 \\
\hline
\end{tabular}
\end{table*}

\begin{figure*}[t]
\centering
\includegraphics[width=0.92\textwidth]{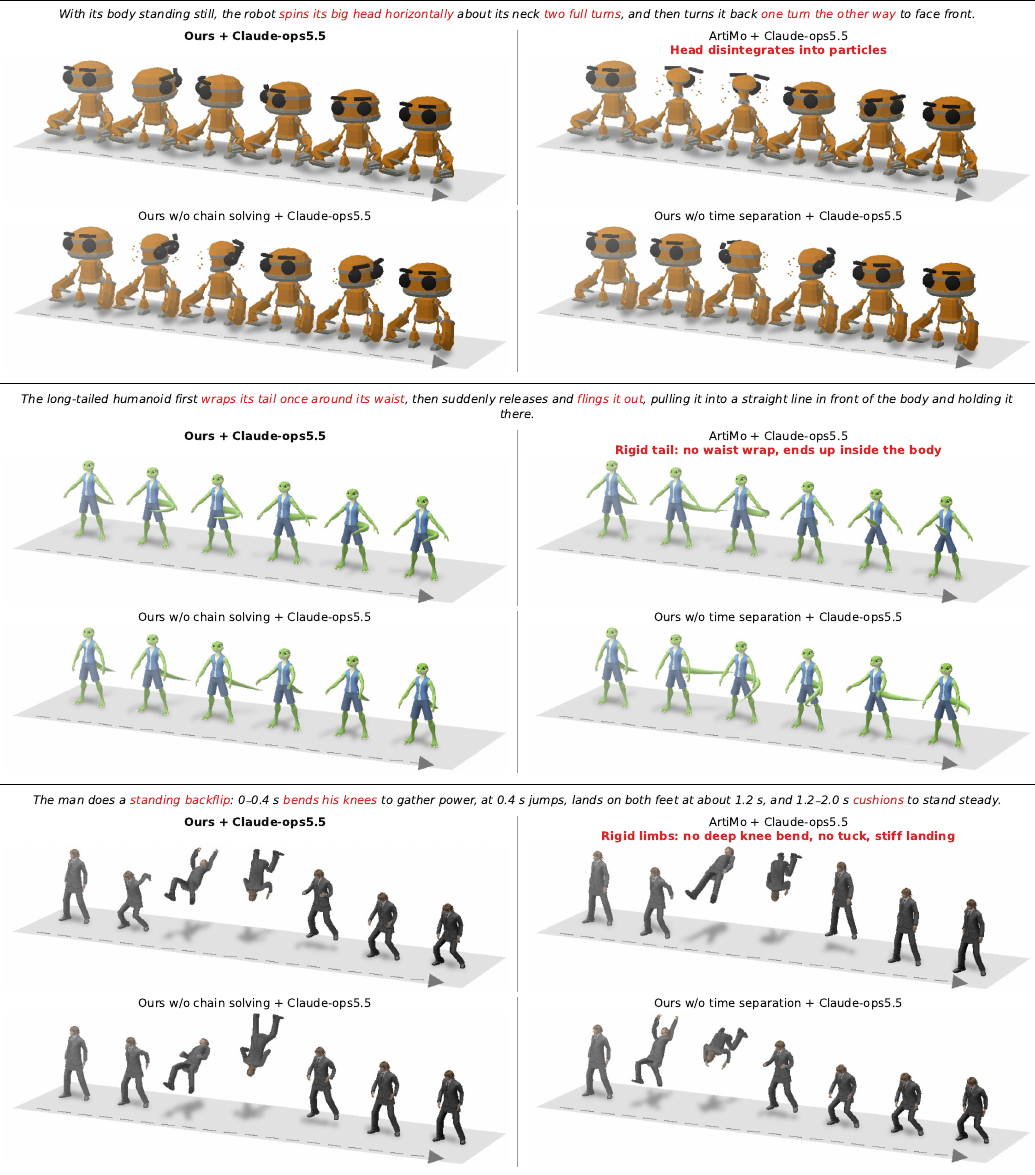}
\caption{Key poses on the robot, long-tailed humanoid, and standing backflip cases under the same assets and instructions. The top row shows Claude-ops5.5 + Ours (left) and ArtiMo with Claude-ops5.5 (right), for comparison with a related agent-based method; the bottom row shows our method without chain solving (left) and without time separation (right), two ablation variants. Red text marks the observed failure of each variant.}
\label{fig:supp_artimo}
\end{figure*}

\begin{figure}[!t]
\centering
\includegraphics[width=\columnwidth]{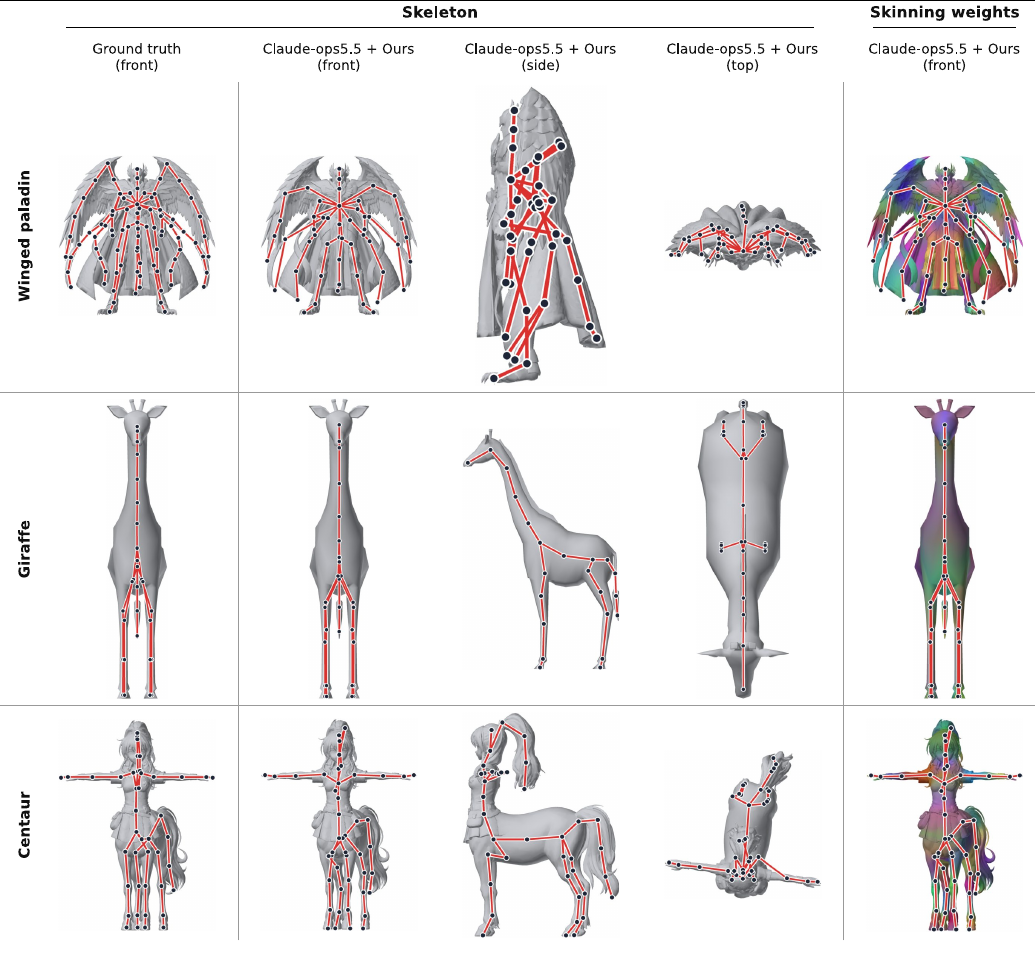}
\caption{Additional rigging results of Claude-ops5.5 + Ours. The ground-truth skeleton is shown beside our predicted skeleton in the front view, followed by the side and top views and the predicted skinning weights.}
\label{fig:supp_rigging}
\end{figure}

\begin{figure*}[!t]
\centering
\includegraphics[width=0.92\textwidth]{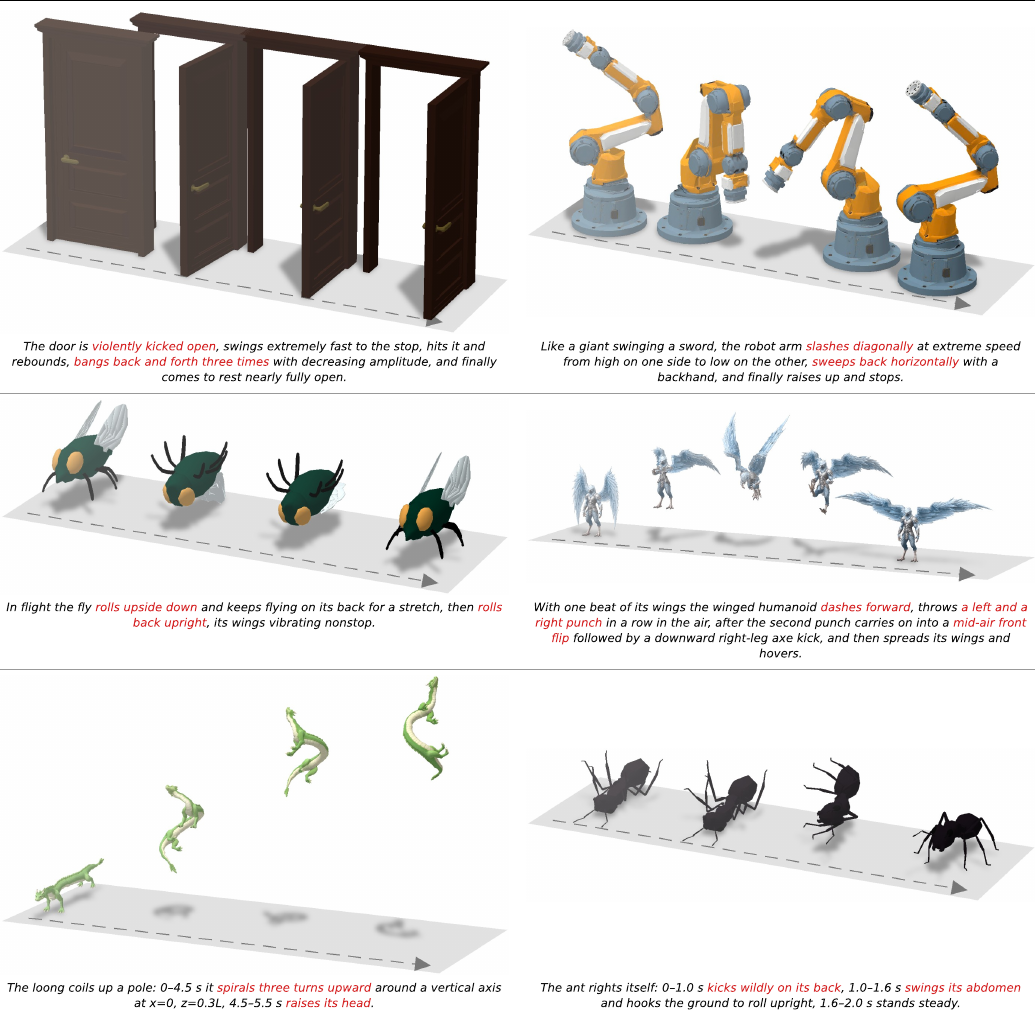}
\caption{Additional key-pose sequences generated by Claude-ops5.5 + Ours on six assets: a door, a robot arm, a fly, a winged humanoid, a loong, and an ant.}
\label{fig:supp_motion}
\end{figure*}


\section{Evaluation Details}
\label{sec:supp_bench}

This section gives the metric definitions and scoring details used in our experiments. Rigging lengths are normalized by $s$, the longest side of the bounding box of the ground-truth rest mesh. Motion lengths are normalized by $L$, the rest-mesh height for text cases and the length fixed by the frozen specification for numeric cases.

\subsection{Baseline Setup and Fairness}
\label{sec:supp_baseline_setup}

As shown in Table~\ref{tab:supp_baseline_setup}, no baseline has access to the benchmark's checking specification or evaluation results during generation; each method only sees the instruction. All methods share the same agent backbone, output format, and asset/rig inputs, and the turn budget is comparable across methods, so score differences reflect the generation pipeline itself rather than unequal access to evaluation signals or unequal optimization effort.

\subsection{Rigging Metrics}
\label{sec:supp_metrics}

\paragraph{Validity and Source Fidelity.}
A submission is valid only if it is a skinned asset that keeps the source mesh. Skinning weights must be nonnegative, sum to one within $10^{-3}$, and use at most four influences per vertex. The skeleton must have a single root and no degenerate bone, and every vertex must be bound. The mesh must keep the vertex count and the triangle set, and every rest position must stay within $10^{-4}s$ of the source, while vertex reordering is allowed. Missing and invalid submissions count as failures, and Source Fidelity is the category-macro average of the valid ratio.

\paragraph{Skeleton and skinning.}
Let $\mathcal P$ and $\mathcal G$ be the predicted and ground-truth joint positions. J2J is the symmetric Chamfer distance and requires no joint correspondence,
\begin{equation}
\mathrm{J2J}=\frac{1}{2s}\Bigl(\frac{1}{|\mathcal P|}\sum_{\mathbf p\in\mathcal P}\min_{\mathbf g\in\mathcal G}\|\mathbf p-\mathbf g\|_2+\frac{1}{|\mathcal G|}\sum_{\mathbf g\in\mathcal G}\min_{\mathbf p\in\mathcal P}\|\mathbf p-\mathbf g\|_2\Bigr).
\label{eq:supp_j2j}
\end{equation}
For mIoU, each vertex is labeled by its dominant bone in both rigs. Area-weighted IoU is computed between predicted and ground-truth parts, parts are matched one to one by the Hungarian algorithm, and unmatched ground-truth parts receive zero before averaging over ground-truth parts. Joint validity is the fraction of predicted joints that lie inside the mesh, with a generalized winding number above $0.5$, or within $0.01s$ of the nearest vertex.

\paragraph{Deformation.}
Each non-leaf joint is rotated by $\pm30^\circ$ about each local axis while the others stay at rest, and the mesh is deformed by linear blend skinning. For each pose, Edge $p_{99}$ takes the 99th percentile of $|l'_e/l_e-1|$ over mesh edges $e$, and the median over poses is reported. Rigid RMS applies to mechanical assets. In each pose, a rigid transform is fitted to every rigid part by the Kabsch algorithm, and the residual RMS divided by $s$ is taken over the worst pose and part. mIoU and joint validity count failures as zero over all submissions, while J2J, Edge $p_{99}$, and Rigid RMS average only over the valid submissions on which they are measurable; all four are then averaged within each category and macro-averaged across categories, as for Source Fidelity.

\subsection{Motion Metrics}
\label{sec:supp_motion_metrics}

\paragraph{Validity.}
Each output is sampled at 30 fps on the reference rig. An output is invalid if it is shorter than the required duration, contains non-finite values, or changes the skeleton or the mesh, and all its applicable dimensions then score zero. In the end-to-end setting~\ref{sec:supp_e2e}, the skeleton and mesh are instead checked against the predicted rig.

\paragraph{Semantic.}
For text cases, the judge scores each rubric item at its allowed levels of full, partial, and zero points, and items involving repetition are scored in proportion to the completed count. A gate checks whether the core action appears, and Semantic and Completeness are zero when the gate fails. A pseudo pass is flagged when global root motion replaces the required part motion, the mesh is rescaled, a required attachment stays static, the motion is almost static, or the root moves more than $0.2L$ in one frame. Such cases score zero in Semantic, Completeness, and Morphology.

\paragraph{Accuracy.}
Each numeric case has a frozen specification of signals, checks, tolerances, and time windows. Signals include positions, velocities, joint and heading angles, distances, heights, and event times, and window boundaries allow a margin of $0.1$\,s. A check with error $e$ and tolerance $\tau$ receives
\begin{equation}
a=\begin{cases}1, & e\le\tau,\\ \operatorname{clip}\left(2-e/\tau,\,0,\,1\right), & e>\tau,\end{cases}
\label{eq:supp_check}
\end{equation}
which decreases linearly from $\tau$ to $2\tau$. A scoring item $k$ worth $n_k$ points receives $n_k\sum_c\omega_c a_c/\sum_c\omega_c$ over its checks $c$, and Accuracy sums the items to 100. Anti-shortcut checks carry no points, but Accuracy becomes zero if any of them fails.

\paragraph{Completeness.}
For text cases, the judge scores whether each phase occurs in the required order and whether sustained requirements hold. Phases share 100 points, or 80 points when 20 points are reserved for sustained requirements. For numeric cases, each phase receives the mean rate of its linked accuracy items, a rate of at least 0.9 earns full points, and phases not covered by the animation score zero.

\paragraph{Morphology and Aesthetics.}
Morphology has four items worth 30, 25, 25, and 20 points, covering correct part use, motion style suited to the morphology, joint ranges, and follow-through of secondary parts. Let $r$ be the fraction of frames in which a joint bent at rest flips its bending direction. The joint-range item then scores zero when $r>0.1$ and at most 12 points when $0.02<r\le0.1$. Aesthetics scores smoothness, weight, timing, follow-through, and pose silhouette, each worth 20 points at the levels 20, 14, 7, and 0.

\paragraph{Quality.}
Quality is computed programmatically. Each term maps its measurement $v$ to $q=\operatorname{clip}\left((z-v)/(z-f),\,0,\,1\right)$, which gives full credit at $f$ and zero at $z$, with the values in Table~\ref{tab:supp_quality}. Term weights depend on the environment of the asset. Terms that do not apply, such as foot skating without ground contact, are removed, and the remaining weights are renormalized to 100.

\paragraph{Overall.}
Overall combines the applicable dimensions $d$ of each case as $\sum_d w_dS_d/\sum_d w_d$, with $w_d$ equal to 0.35 for Semantic or Accuracy, 0.20 for Completeness, 0.15 for Morphology, 0.20 for Quality, and 0.10 for Aesthetics. Text and numeric cases use the same weights.

\paragraph{Judging.}
The judge views time-stamped front and side contact sheets of the exported animation together with the instruction, the rubric, and a short summary of the root trajectory, and scores each item once. Every score must be one of the allowed levels and cite the view, the time, and the observed behavior. Quantities computed by the program are never judged visually.

\subsection{User Study Setup}
\label{sec:supp_user_study}
The main-text user study invites 30 volunteers to score outputs of GPT6.1-sol + UniMate, Seedance 2.5 + Claude-ops5.5, and Claude-ops5.5 + Ours, with the setup given in Table~\ref{tab:supp_user_study}. Text cases are scored on Semantic, Completeness, Morphology, and Aesthetics, while numeric cases are scored only on Morphology and Aesthetics, using the same 0--100 scale as the main-text rubric.

\paragraph{Ethics.}
This study is a low-risk anonymized rating task that involves no sensitive personal information, and was exempt from formal ethics review under institutional policy. Participants gave informed consent before rating, and method identities were anonymized as A/B/C during scoring. Incidentally, all benchmark assets come from open-source sources, and the annotators who labeled the data were paid for their work.

\section{Framework Implementation Details}
\label{sec:supp_agent}

This section details the deterministic executor used for rigging, motion solving, and mesh-feedback refinement. We use the main-text notation: $M=(V,F)$ is the input mesh, $S$ the rigging configuration, $R$ the rig, and $P$ the motion program.

\subsection{Rigging Details}
\label{sec:supp_agent_rigging}

\paragraph{Joint reconstruction and topology.}
For joint $j$, the agent annotates pixel position $\mathbf p_{jv}$ and confidence $\gamma_{jv}$ in calibrated view $v$. Let $\mathcal V_j$ be its annotated views and $A_v\mathbf x+\mathbf b_v$ the orthographic projection. The executor reconstructs the joint by weighted least squares:
\begin{equation}
\widehat{\mathbf j}_j
:=\underset{\mathbf x\in\mathbb R^3}{\arg\min}
\sum_{v\in\mathcal V_j}\gamma_{jv}
\left\|A_v\mathbf x+\mathbf b_v-\mathbf p_{jv}\right\|_2^2.
\label{eq:agent_joint_lifting}
\end{equation}
For a joint annotated in only one view, ray--mesh intersections determine depth. Ordered bone chains and attachment joints define parent relationships $\operatorname{pa}(j)$. The root controls global translation and orientation and is typically placed at the trunk base connecting the main chains, rather than fixed to a human pelvis. In an existing rig, it is the unique joint whose parent lies outside the skin-joint set. Part mappings record chains, categories, and laterality to associate instruction terms with actual joints.

\paragraph{Geodesic-gated skinning.}
Let $d_{ij}$ be the Euclidean distance from vertex $\mathbf v_i$ to the bone segments driven by joint $j$, and $g_{ij}$ the bounded geodesic distance along the mesh surface. With gating parameters $\rho$ and $\delta$, the allowed influence set is $\Omega_i=\{j:a_{ij}=1\}$. The executor initializes weights using a distance-decay kernel $K$ and smooths them over surface neighbors:
\begin{equation}
\begin{aligned}
a_{ij}&=\mathbf 1\!\left[g_{ij}\leq\rho d_{ij}+\delta\right],\\
\mathbf w_i^{(0)}&=\Pi_{\Omega_i}\!\left(K(\mathbf d_i)\right),\\
\mathbf w_i^{(m+1)}&=\Pi_{\Omega_i}\!\left((1-\eta)\mathbf w_i^{(m)}+\eta\,\overline{\mathbf w}_i^{(m)}\right).
\end{aligned}
\label{eq:agent_skinning}
\end{equation}
Here, $\mathbf d_i$ collects vertex-to-bone distances, $\eta$ is the smoothing coefficient, and $\overline{\mathbf w}_i^{(m)}$ is the neighborhood mean at iteration $m$. The operator $\Pi_{\Omega_i}$ restricts weights to allowed influences, retains at most four, and normalizes them. Geodesic gating limits weight transfer between spatially adjacent but surface-separated appendages. The agent can further restrict influences by mesh part or view region and bind rigid parts entirely to individual bones. The exported rig is reloaded and tested with probe deformations.

\subsection{Motion Control and Kinematic Solving}
\label{sec:supp_agent_motion}

\paragraph{Control parameters.}
In the main-text motion program $P$, root control $\mathcal R_0=(j_0,o_0,\kappa_0)$ acts on the hierarchy root $j_0$ of $R$. Each part program $P_p=(\Gamma_p,o_p,\kappa_p,\mathcal W_p)$ specifies a joint chain, an operator, its parameters, and contact windows. Operators support end-effector targets, rotation, aiming, and translation. Parameters include control curves, reference frames, joint ranges, and bend-direction preferences. Time can be specified in seconds, as a fraction of duration, or by event and phase references. Repetition is controlled by count or frequency within a phase.

\paragraph{Root trajectory.}
For curve-based root control, $\kappa_0$ contains displacement curves $x_0,y_0,z_0$ and yaw, with optional pitch, roll, or forward/lateral speed. Let $\ell$ be a characteristic lower-limb length, or the skeleton scale when no lower limbs are present, and $Q_0$ the body frame in the rest pose. The basic displacement term gives the root's world position:
\begin{equation}
\mathbf t_0(t)=\mathbf t_0^{\mathrm{rest}}
+\ell Q_0\begin{bmatrix}x_0(t)\\y_0(t)\\z_0(t)\end{bmatrix}.
\label{eq:agent_root_motion}
\end{equation}
Displacement curves use $\ell$ as their unit, so metric targets are converted to this scale. Speed curves add integrated displacement along the current yaw heading, and a configured rotation pivot adds compensation. The executor converts the target position and orientation into the parent-relative local transform $L_{j_0}(t)$. Forward kinematics (FK) propagates this transform through the hierarchy rather than planning the trajectory.

\paragraph{End-effector solving.}
At each sampled frame, the executor establishes the root pose and preceding local rotations before solving end-effector targets with inverse kinematics (IK). Contact windows anchor targets in world space unless sliding is specified.

The end-effector operator handles chains of arbitrary length, such as legs, tails, tentacles, and spines. For part $p$, the program selects a contiguous chain $\Gamma_p=(j_0,\dots,j_N)$ with $N\ge2$ and specifies only the trajectory $\mathbf y_n$ of the end joint $j_N$ through keys, plants, or gait windows. The interior joints $j_1,\dots,j_{N-1}$ receive no individual targets. Instead, they share one scalar bend $\theta$, which is distributed by the bend side $\boldsymbol\pi$, the flexion limits $[\beta_k^-,\beta_k^+]$, and the bend weights $w_k$ in $\kappa_p$, normalized so that $\sum_k|w_k|=1$. Let $\mathbf x_k$ be the joint positions after the root and preceding rotations, $\mathbf b_k=\mathbf x_{k+1}-\mathbf x_k$ the current bones, and $\mathbf h$ the unit hinge axis along $\boldsymbol\pi\times\sum_k\mathbf b_k$. Keeping the base fixed, the executor bends the chain about $\mathbf h$ and rebuilds it bone by bone:
\begin{equation}
\begin{aligned}
\beta_k(\theta)&=\operatorname{clip}\!\left(\theta w_k,\beta_k^-,\beta_k^+\right),\\
\varphi_k(\theta)&=\textstyle\sum_{m=1}^{k}\beta_m(\theta),\quad \varphi_0=0,\quad \widetilde{\mathbf x}_0=\mathbf 0,\\
\widetilde{\mathbf x}_{k+1}(\theta)&=\widetilde{\mathbf x}_k(\theta)+\operatorname{Rot}\!\left(\mathbf h,\varphi_k(\theta)\right)\mathbf b_k,\\
\theta_n^*&=\underset{\theta\in[\theta^-,\theta^+]}{\arg\min}\,
\frac{\bigl|\|\widetilde{\mathbf x}_N(\theta)\|_2-d_n\bigr|}{d_n}\\
&\quad+\mu\left|\theta-\theta_{n-1}^*\right|.
\end{aligned}
\label{eq:supp_chain_bend}
\end{equation}
Here, $d_n=\|\mathbf y_n-\mathbf x_0\|_2$ is the target distance at frame $n$, $[\theta^-,\theta^+]$ is the range of total bend admitted by the flexion limits, and $\mu$ is a small continuity weight that suppresses frame-to-frame flips. The minimizer is found by a grid search followed by bisection. The bent chain is then rotated rigidly about the base so that its tip points at the target and its bend side faces $\boldsymbol\pi$:
\begin{equation}
\begin{aligned}
A_n&=F\!\left(\mathbf y_n-\mathbf x_0,\boldsymbol\pi\right)F\!\left(\widetilde{\mathbf x}_N,\widetilde{\mathbf x}_N\times\mathbf h\right)^{\top},\\
\mathbf x_k'&=\mathbf x_0+A_n\widetilde{\mathbf x}_k(\theta_n^*),\\
R_{j_k}'&=A_n\operatorname{Rot}\!\left(\mathbf h,\varphi_k(\theta_n^*)\right)R_{j_k},\quad 1\le k<N.
\end{aligned}
\label{eq:supp_chain_align}
\end{equation}
Here, $F(\mathbf a,\mathbf c)$ is the orthonormal frame whose first axis is $\mathbf a$ and whose second axis is the component of $\mathbf c$ orthogonal to $\mathbf a$, and $R_{j_k}$ is the world rotation of $j_k$ before solving.

\paragraph{Relative motion of chain joints.}
Each interior joint thus moves only through its neighbors. Since $\mathbf x_{k+1}'-\mathbf x_k'=A_n\operatorname{Rot}(\mathbf h,\varphi_k)\mathbf b_k$, every joint keeps its bone length to both the parent and the child, rotates by the bend accumulated over the joints above it, and preserves the twist and non-planar shape of the current chain. A positive weight bends a joint toward $\boldsymbol\pi$, a negative weight bends it the other way, which yields alternating folds such as a Z-shaped foreleg, and a zero weight keeps the joint straight. The base $j_0$ is not moved by this operator and follows its parent. Joints below $j_N$ follow through FK. When the chain carries a rigid tip segment with a specified end orientation, the target is first offset by that segment under the orientation, and the orientation of $j_N$ is written after solving. Chains are solved in hierarchy order, so a chain attached to a previously solved chain starts from the updated base.

\paragraph{Unreachable targets.}
If $d_n$ lies outside the reach admitted by the flexion limits, $\theta_n^*$ gives the closest reachable distance and $A_n$ still aligns the tip with the target direction. The end joint therefore stops at $\mathbf x_0+\|\widetilde{\mathbf x}_N(\theta_n^*)\|_2\,(\mathbf y_n-\mathbf x_0)/d_n$, on the line from the base toward the target. The residual $\|\mathbf x_N'-\mathbf y_n\|_2$ is measured against the original target and enters the IK term $e_{\mathrm{IK}}$ of Eq.~\eqref{eq:agent_refine_score}. This closest-reachable guarantee holds within the one-parameter bend family of Eq.~\eqref{eq:supp_chain_bend} rather than over the full IK search space, and the continuity term $\mu|\theta-\theta_{n-1}^*|$ trades a small amount of this optimality for frame-to-frame stability. When $d_n=0$ or $\boldsymbol\pi$ is parallel to the chain direction, the hinge axis $\mathbf h$ is undefined; the executor then falls back to the hinge axis and alignment frame from the previous frame, or to a fixed default axis on the first frame.

\paragraph{Arm arcs.}
For the arm-arc operator, parameters $\kappa_{p,n}$ at frame $n$ construct a target from angle, radius, and lateral offset. Previously computed parent transforms $G^{\mathrm{pre}}(t_n)$ provide the reference-frame conversion. Shoulder flexion, abduction, swivel, and elbow flexion angles $\boldsymbol\alpha$ are solved as
\begin{equation}
\begin{aligned}
\mathbf y_{p,n}&=\operatorname{Target}(\kappa_{p,n};R,G^{\mathrm{pre}}(t_n)),\\
\mathbf e_p(\boldsymbol\alpha)&=l_{p,1}\mathbf n_1(\boldsymbol\alpha)+l_{p,2}\mathbf n_2(\boldsymbol\alpha),\\
E_{p,n}(\boldsymbol\alpha)&=\left\|\frac{\mathbf e_p(\boldsymbol\alpha)-\mathbf y_{p,n}}{l_{p,1}+l_{p,2}}\right\|_2^2\\
&\quad+\lambda_{a,p}^2\left(\alpha_{\mathrm{sw}}-\overline\alpha_{p,n}\right)^2,\\
\boldsymbol\alpha_{p,n}^*&=\underset{\boldsymbol\alpha\in\mathcal D_p}{\arg\min}\,E_{p,n}(\boldsymbol\alpha).
\end{aligned}
\label{eq:agent_arm_ik}
\end{equation}
Here, $l_{p,1},l_{p,2}$ are the upper- and lower-arm lengths in $R$, $\mathbf n_1,\mathbf n_2$ are bone directions in the parent-driven chest frame, and $\mathbf y_{p,n}$ is the shoulder-relative target after workspace correction. The joint-angle domain $\mathcal D_p$, regularization coefficient $\lambda_{a,p}$, and elbow-plane preference $\overline\alpha_{p,n}$ are specified by $\kappa_p$, not additional per-frame controls.

\paragraph{Pose propagation and mesh deformation.}
Given local rigid transforms $L_j(t)$, FK computes global transforms, and linear blend skinning (LBS) deforms the mesh:
\begin{equation}
\begin{aligned}
G_j(t)&=G_{\operatorname{pa}(j)}(t)\,L_j(t),\\
\widetilde{\mathbf v}_i(t)&=\left[\sum_{j\in\mathcal J}w_{ij}\,G_j(t)B_j^{-1}\overline{\mathbf v}_i\right]_{1:3}.
\end{aligned}
\label{eq:agent_fk_lbs}
\end{equation}
Here, $\mathcal J$ is the skin-joint set, $B_j^{-1}$ an inverse bind matrix, and $\overline{\mathbf v}_i$ a homogeneous vertex coordinate. FK preserves non-skin scene ancestors and uses an identity parent transform only for nodes without a parent. Root motion propagates through the hierarchy, while IK adjusts contact limbs toward anchored targets. The exported animated GLB is validated by reloading it.

\subsection{Mesh-feedback Refinement}
\label{sec:supp_agent_refinement}

Refinement makes incremental parameter adjustments on top of $P$. Using the deformed mesh, the executor accumulates four additive correction terms across iterations and layers them onto the kinematic solve, instead of editing $\kappa_0$ or any $\kappa_p$ directly. The four terms are an arm-clearance offset on arm-arc targets, a foot lift on stance height, an end-effector offset on collision-avoiding effector targets, and a root-height lift on top of the position $\mathcal R_0$ specifies. Each term uses the corresponding operator to apply an incremental adjustment on its own parameters; the arm-clearance offset perturbs targets through operator $o_p=\mathrm{arm\_arc}$, the foot lift and end-effector offset perturb targets through operator $o_p\in\{\mathrm{effector},\mathrm{foot\_swing},\mathrm{contact\_path}\}$ with the end-effector offset zeroed during each part's own $\mathcal W_p$ so planted frames stay anchored, and the root-height lift perturbs the height of the position through root operator $o_0$. Candidate $r$ is evaluated by
\begin{equation}
\begin{aligned}
\sigma_r={}&e_{\mathrm{env}}^{(r)}
+\frac{\lambda_{\mathrm{IK}}e_{\mathrm{IK}}^{(r)}+\lambda_g h_g^{(r)}}{\ell}\\
&+\frac{\chi_c h_c^{(r)}+\chi_t\lambda_g e_t^{(r)}}{\ell}.
\end{aligned}
\label{eq:agent_refine_score}
\end{equation}
Here, $e_{\mathrm{env}}$ measures forearm intrusion into the trunk envelope, $e_{\mathrm{IK}}$ is the IK residual, and $h_g$ is ground-penetration depth. Optional collision depth $h_c$ and ground-touch error $e_t$ are enabled by $\chi_c$ and $\chi_t$. The coefficients $\lambda_{\mathrm{IK}}$ and $\lambda_g$ weight the residual terms, and $\ell$ is the characteristic length defined above. A candidate is accepted only if the score strictly decreases. Otherwise, refinement stops and returns the best accepted result, with at most five re-solves. $\sigma_r$ is computed solely from the current asset's own mesh and skeleton, the program $P$ derived from the instruction, and this geometric feedback; it never reads a hidden grading rubric, a benchmark's checking specification, or any evaluation result.

This additive correction loop runs inside a single executor call and only covers small, continuous deviations such as residual interpenetration or ground contact. If the exported animation is still clearly unnatural, e.g., an end-effector chain $\Gamma_p$ is poorly chosen for the required reach or a targeted part still collides severely, the agent inspects the result and can revise $P$ itself, for example reselecting $\Gamma_p$, switching $o_p$, or rewriting $\kappa_p$, then invokes the executor again in a new turn.

\section{Additional Results}
\label{sec:supp_results}
\subsection{End-to-end evaluation}
\label{sec:supp_e2e}
The main paper evaluates motion on a provided reference rig so that motion scores are not entangled with rigging errors. Here we remove this separation, so the input is only the raw, unrigged mesh together with the motion instruction, and Claude-ops5.5 + Ours automatically runs the full pipeline from rigging to motion generation, after which only the resulting motion is scored. We evaluate on the humanoid, axial, and mixed-limb categories, which differ substantially in skeletal topology and body shape and are chosen to be representative. The same motion instructions and scoring are used, and only the rig source differs. Predicted rigs still yield solid results, with Overall scores of 49.80, 34.58, and 51.20 on the three categories, and Quality and Aesthetics stay close to the values obtained with the reference rigs. On mixed-limb assets, Overall (48.33$\to$51.20) is even slightly higher than with the reference rig, despite a drop in Accuracy (48.39$\to$37.51). Our analysis suggests this relates to mixed-limb assets typically having outstretched limbs with little complex clothing or curling, a body shape that is generally easier to rig accurately, and also because this is an end-to-end pipeline, rigging and motion generation are directly coupled, and this coupling may itself contribute to the Overall gain. These results suggest that our agent framework can carry out reasonably reliable end-to-end motion generation from a raw mesh.

\subsection{Comparison with Related Agent-based Methods}
\label{sec:supp_artimo}
ArtiMo~\cite{zou2026artimo} is a fairly representative related agent-based motion generation method, but its articulation model is more restricted than ours, and it is not specifically designed to handle diverse or complex morphologies. Fig.~\ref{fig:supp_artimo} (top-left and top-right of each case) compares it with Claude-ops5.5 + Ours on the same assets and instructions, with ArtiMo driven by Claude-ops5.5. On the robot, both spin the head about the neck, but ArtiMo's head breaks into particles during the turns, whereas ours stays intact and returns to face front. For the long-tailed humanoid, ours wraps the tail around the waist and then flings it into a straight line in front of the body, while ArtiMo's tail stays rigid, never wraps the waist, and ends up inside the body. For the standing backflip, ours bends the knees, tucks in the air, and cushions the landing, whereas ArtiMo's limbs stay rigid, with no deep knee bend or tuck and a stiff landing. These cases require the motion of multi-joint parts such as a tail or limbs to follow the phases of the instruction, consistent with the more restricted articulation model of ArtiMo, while our programs can set targets for each chain in each phase.

\subsection{Ablation Study}
\label{sec:supp_artimo_ablation}
Fig.~\ref{fig:supp_artimo} (bottom-left and bottom-right of each case) further shows two ablations, Ours w/o chain solving and Ours w/o time separation, both compared against the full method at the top-left of each case.

Chain solving distributes an end-effector target across the intermediate joints of the whole chain. Without it, the robot's head likewise disintegrates briefly into particles during the turn, the long-tailed humanoid's tail stays in a near-rest, drooping shape throughout, neither wrapping the waist nor flinging out, and the legs adopt an unnatural, overly raised pose during the crouch phase of the standing backflip. These observations indicate that without solving the chain as a whole, multi-joint parts struggle to reach the target shape required by the instruction and tend to show stiff or disintegrated intermediate poses.

Time separation splits an instruction into an ordered sequence of phases that are then solved separately. Without it, the long-tailed humanoid's tail alternates between an extended and a drooping shape across frames, flinging toward the front before it has wrapped the waist, and the frame corresponding to the power-gathering phase of the standing backflip shows the arms thrown overhead, an action that does not belong to that phase. These observations indicate that without separating phases, the target poses may still be reached, but their order can be inconsistent with the phase structure of the instruction.

\subsection{Qualitative results}
\label{sec:supp_qual}
To further demonstrate the effectiveness of our framework, we provide more visualizations of its rigging and motion generation results in Fig.~\ref{fig:supp_rigging} and Fig.~\ref{fig:supp_motion}.

Fig.~\ref{fig:supp_rigging} shows rigging results of Claude-ops5.5 + Ours on a winged paladin, a giraffe, and a centaur, with the ground-truth skeleton placed beside our skeleton in the front view. In all three cases, the predicted skeleton matches the ground truth in overall topology and joint count. On the paladin, the skeleton follows the ground truth with a spine, two arms, two legs, and a long chain through each wing, so the wings and the cape can move independently of the torso, though the hip and shoulder joints sit slightly lower than in the ground truth. On the giraffe, the neck chain, the four legs, and the tail match the layout of the ground truth, and the side and top views confirm that the joints lie inside the body, though the forelimb joints leave some room to better match the ground truth. On the centaur, our skeleton correctly joins a humanoid upper body (spine and two arms) with a quadruped lower body (four horse legs and a tail) in a single chain, and the side view confirms that the torso and limb orientations match the ground truth, with a similar margin for improvement in the forelimb joints. The skinning weights follow the same structure. They vary smoothly along the neck and legs of the giraffe, and separate the humanoid upper body from the quadruped lower body of the centaur into distinct regions.

Fig.~\ref{fig:supp_motion} shows key poses on six assets from different morphology categories. The mechanical assets follow rigid and numerically specified motion, with the door swinging open and bouncing back, and the robot arm slashing and sweeping. The fly rolls onto its back and returns upright, and the winged humanoid dashes, strikes, flips, and hovers, which shows that the framework handles winged assets and instructions with many ordered phases. The loong spirals upward along a vertical axis, and the ant kicks on its back before rolling upright, covering an axial body and a multi-legged body. In each case the poses follow the order of phases in the instruction.

\end{document}